\documentclass[runningheads]{llncs}

\usepackage{amsfonts}
\usepackage{booktabs}
\usepackage{csquotes}
\usepackage{mathtools}
\usepackage{multirow}
\usepackage{stmaryrd}
\usepackage{tablefootnote}
\usepackage[detect-weight=true,binary-units=true]{siunitx}
\usepackage{xfp}

\newcommand{\pc}[1]{\SI[round-mode=places,round-precision=1]{#1}{\percent}}

\usepackage{algorithm}
\usepackage[noend]{algpseudocode}

\algtext*{EndWhile}

\usepackage{hyperref}
\hypersetup{
pdftitle={Regularization in Spider-Style Strategy Discovery and Schedule Construction},
pdfauthor={Filip Bártek, Karel Chvalovský, Martin Suda},
pdfkeywords={saturation-based theorem proving, proving strategies, strategy schedule construction, Vampire}
}
\hypersetup{hidelinks}

\usepackage{glossaries}

\newacronym{asp}{ASP}{answer set programming}
\newacronym{atp}{ATP}{automatic theorem prover}
\newacronym{cpu}{CPU}{central processing unit}
\glsunset{cpu}
\newacronym{fol}{FOL}{first-order logic}
\newacronym{mi}{Mi}{megainstruction}
\newacronym{par}{PAR}{penalized average runtime}
\newacronym{pcs}{PCS}{parameter configuration space}
\newacronym{sat}{SAT}{propositional satisfiability}
\newacronym{smt}{SMT}{satisfiability modulo theories}
\newacronym{tptp}{TPTP}{Thousands of Problems for Theorem Provers}
\newacronym{ucb}{UCB}{upper confidence bound}
\newacronym{vbs}{VBS}{virtual best solver}

\newacronym{ciirc}{CIIRC}{Czech Institute of Informatics, Robotics and Cybernetics}
\newacronym{ctu}{CTU}{Czech Technical University in Prague}
\newacronym{fel}{FEL}{Faculty of Electrical Engineering}

\newacronym{casc}{CASC}{CADE ATP System Competition}
\newacronym{casc-14}{CASC-14}{CADE-14 ATP System Competition}
\newacronym{casc-j11}{CASC-J11}{11th IJCAR ATP System Competition \cite{DBLP:journals/aicom/SutcliffeD23}}
\newacronym{tfa}{TFA}{Typed (monomorphic) First-order with Arithmetic theorems}
\newacronym{fof}{FOF}{First-Order Form theorems}
\newacronym{fnt}{FNT}{First-order form Non-Theorems}
\newacronym{ueq}{UEQ}{Unit EQuality clause normal form theorems}

\glsdisablehyper

\usepackage{textcomp}

\usepackage[capitalise]{cleveref}
\crefname{figure}{Fig.}{Figures}
\crefname{section}{Sect.}{Sections}

\newcommand{\mli}[1]{\mathit{#1}}

\newcommand{\Instance}{p}

\DeclareMathOperator{\argmax}{argmax}

\newcommand{\Nat}{\mathbb{N}}

\newcommand{\Mi}[1]{\SI{#1}{\acrshort{mi}}}

\newcommand{\SolveTime}{\ensuremath{E}}

\newcommand{\SolveTimeP}[2]{\SolveTime(#1, #2)}

\newcommand{\Acknowledgments}{This work was supported by
	the Czech Science Foundation grants 20-06390Y and 24-12759S,
	the European Regional Development Fund under the Czech project AI\&Reasoning no.~CZ.02.1.01/0.0/0.0/15\_003/0000466,
	the project RICAIP no.~857306 under the EU-H2020 programme, and
	the Grant Agency of the Czech Technical University in Prague, grant no.~SGS20/215/OHK3/3T/37.}

\usepackage{bbding}
\usepackage{todonotes}

\presetkeys{todonotes}{disable}{}

\usepackage{etoolbox}

\newtoggle{PREPRINT}
\toggletrue{PREPRINT} 

\newtoggle{BUDGETS}
\toggletrue{BUDGETS}

\iftoggle{PREPRINT}{
\toggletrue{BUDGETS}
}{}

\title{Regularization in Spider-Style Strategy Discovery and Schedule Construction\iftoggle{PREPRINT}{\thanks{This version of the contribution has been accepted for publication, after peer review but is not the Version of Record and does not reflect post-acceptance improvements, or any corrections. The Version of Record is available online at: \doi{10.1007/978-3-031-63498-7_12}}}{\thanks{Manuscript with additional appendices \cite{bartek2024regularization}: \url{https://arxiv.org/abs/2403.12869}}}}

\def\orcidID#1{\href{http://orcid.org/#1}{\raisebox{-1.25pt}{\includegraphics{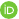}}}}

\author{Filip Bártek\inst{1,2}\Envelope\orcidID{0000-0002-1822-2651} \and
Karel Chvalovský\inst{1}\orcidID{0000-0002-0541-3889} \and 
Martin Suda\inst{1}\orcidID{0000-0003-0989-5800}}
\authorrunning{F. Bártek et al.}

\institute{
\href{https://www.ciirc.cvut.cz/}{\acrlong{ciirc}}\\
\href{https://www.cvut.cz/}{\acrlong{ctu}}, Prague, Czech Republic\\
\and
\href{https://fel.cvut.cz/}{\acrlong{fel}}\\
\href{https://www.cvut.cz/}{\acrlong{ctu}}, Prague, Czech Republic\\
\href{mailto:filip.bartek@cvut.cz,karel.chvalovsky@cvut.cz,martin.suda@cvut.cz}{\email{\{filip.bartek,karel.chvalovsky,martin.suda\}@cvut.cz}}\\
}

\begin{document}

\maketitle

\begin{abstract}

%

To achieve the best performance, automatic theorem provers often rely on schedules of diverse proving strategies to be tried out (either sequentially or in parallel) on a given problem. 
In this paper, we report on a large-scale experiment with discovering strategies for the Vampire prover, targeting the FOF fragment of the TPTP library and constructing a schedule for it, based on the ideas of Andrei Voronkov's system Spider. 
We examine the process from various angles, discuss the difficulty (or ease) of obtaining a strong Vampire schedule for the CASC competition, and establish how well a schedule can be expected to generalize to unseen problems and what factors influence this property.

\end{abstract}


\newcommand{\Problem}{\Instance}

\section{Introduction}

In 1997 at the CADE conference, the automatic theorem prover Gandalf \cite{DBLP:journals/jar/Tammet97}
surprised its contenders at the CASC-14 competition \cite{SS98} and won the MIX division there. 
One of the main innovations later identified as a key to Gandalf's success was the use of multiple 
theorem proving strategies executed sequentially in a time-slicing fashion 
\cite{GandalfDescriptionCASC-15,DBLP:conf/cade/Tammet98}. Nowadays, it is well accepted that 
a single, universal strategy of an Automatic Theorem Prover (ATP) is invariably inferior, in terms of performance,
to a well-chosen portfolio of complementary strategies, most of which do not even need to be complete or very strong in isolation.


Many tools have already been designed to help theorem prover developers discover new proving strategies 
and/or to combine them and construct proving schedules \cite{DBLP:journals/jar/KuhlweinU15,DBLP:conf/gcai/Urban15,DBLP:conf/gcai/SchaferS15,DBLP:conf/cpp/JakubuvU17,DBLP:conf/mkm/HoldenK21,DBLP:conf/mkm/HulaJJK22,DBLP:conf/paar/Schurr22,Spider}. For example, Sch{\"{a}}fer and Schulz employed genetic algorithms \cite{DBLP:conf/gcai/SchaferS15} for the invention of strong strategies for the E prover \cite{DBLP:conf/cade/0001CV19}, 
Urban developed BliStr and used it to significantly strengthen strategies for the same prover 
via iterative local improvement and problem clustering \cite{DBLP:conf/gcai/Urban15}, and, more recently, 
Holden and Korovin applied similar ideas in their HOS-ML system \cite{DBLP:conf/mkm/HoldenK21} to produce 
schedules for iProver \cite{DBLP:conf/cade/Korovin08}. The last work mentioned---as well as, e.g., 
MaLeS \cite{DBLP:journals/jar/KuhlweinU15}---also include a component for \emph{strategy 
selection}, the task of predicting, based on the input problem's features, which strategy 
will most likely succeed on it.
(Selection is an interesting topic, which is, however, orthogonal to our work and will not be further discussed here.)

For the Vampire prover \cite{DBLP:conf/cav/KovacsV13}, schedules were for a long time constructed by Andrei Voronkov
using a tool called Spider, about which little was known until recently. 
Its author finally revealed the architectural building blocks of Spider and the ideas behind them
at the Vampire Workshop 2023, declaring Spider
``a secret weapon behind Vampire's success at the CASC competitions''
and ``probably the most useful tool for Vampire's support and development'' \cite{Spider}.
Acknowledging the importance of strategies for practical ATP usability,
we decided to analyze this powerful technology on our own.

	

In this paper, we report on a large-scale experiment with discovering strategies for Vampire,
based on the ideas of Spider (recalled in Sect.~\ref{sec:spider}).\footnote{Not claiming any credit for these, potential errors in the explanation are ours alone.}
We target the FOF fragment of the TPTP library \cite{Sutcliffe2017}, probably the most comprehensive benchmark set available for first-order theorem proving.
As detailed in Sect.~\ref{sect:discover}, we discover and evaluate (on all the FOF problems) more than \num{1000} targeted strategies to serve as building blocks 
for subsequent schedule construction.



Research on proving strategies is sometimes frowned upon as mere ``tuning for competitions''.
While we briefly pause to discuss this aspect in Sect.~\ref{sec:casc}, 
our main interest in this work is to establish how well a schedule can be expected to generalize to unseen problems.
For this purpose, we adopt the standard practice from statistics to randomly split the available problems
into a train set and a test set, construct a schedule on one, and evaluate it on the other.
In Sect.~\ref{sec:regul}, we then identify several techniques that \emph{regularize}, i.e., 
have the tendency to improve the test performance while possibly sacrificing the training one.

Optimal schedule construction under some time budget can be expressed as a mixed integer program
and solved (given enough time) using a dedicated tool \cite{DBLP:conf/mkm/HoldenK21,DBLP:conf/paar/Schurr22}.
Here, we propose to instead use a simple heuristic from the related set cover problem \cite{DBLP:journals/mor/Chvatal79},
which leads to a polynomial-time greedy algorithm (Sect.~\ref{sec:greedy_enters_the_scene}).
The algorithm maintains the important ability to assign different time limits to different strategies,
is much faster than optimal solving (which may overfit to the train set in some scenarios), 
allows for easy experimentation with regularization techniques,
and, in a certain sense made precise later, does not require committing to a single predetermined time budget.

In summary, we make the following main contributions:
\begin{itemize}


\item
We outline a pragmatic approach to schedule construction that uses a greedy algorithm (\cref{sec:greedy}),
contrasting it with optimal schedules in terms of the quality of the schedules and the computational resources required for their construction (\cref{sec:experimental}).
In particular, our findings demonstrate a relative efficacy of the greedy approach for datasets similar to our own.

\item
Leveraging the adaptability of the greedy algorithm,
we introduce a range of regularization techniques aimed at improving the robustness of the schedules in unseen data (\cref{sec:regul}).
To the best of our knowledge, this represents the first systematic exploration into regularization of strategy schedules.
%
\item The strategy discovery and evaluation is a computationally expensive process, which in our case took 
more than twenty days on 120 CPU cores. At the same time, 
there are further interesting questions concerning Vampire's strategies than we could answer in this work. 
To facilitate research on this paper's topic, we made
the corresponding data set available online \cite{bartek_2024_10814478}.

\end{itemize}


\section{Preliminaries}


\label{sec:prelim}

The behavior of Vampire is controlled by approximately one hundred \emph{options}. 
These options configure the preprocessing and clausification steps, 
control the saturation algorithm, clause and literal selection heuristics, determine
the choice of generating inferences as well as redundancy elimination and simplification rules, and more. 
Most of these options range over the Boolean or a small finite domain, a few are numeric (integer or float), 
and several represent ratios.

Every option has a \emph{default} value, which is typically the most universally useful one.
Some option settings make Vampire incomplete. This is automatically recognized,
so that when the prover finitely saturates the input without discovering a contradiction,
it will report ``unknown'' (rather than ``satisfiable'').

A \emph{strategy} is determined by specifying the values of all options. 
A \emph{schedule} is a sequence $(s_i,t_i)_{i=1}^n$ of strategies $s_i$ together with
assigned time limits $t_i$, intended to be executed in the prescribed order.
We stress that in this work we do not consider schedules that would branch depending on problem features.


\subsection{Spider-Style Strategy Discovery and Schedule Construction}
\label{sec:spider}





We are given a set of problems $P$ and a prover with its space of available strategies $\mathbb{S}$.
\emph{Strategy discovery} and \emph{schedule construction} are two separate phases. 
We work under the premise that the
larger and more diverse a set of strategies we first collect, the better for later constructing a good schedule.


Strategy discovery consists of three stages: random probing, strategy optimization, and evaluation,
which can be repeated as long as progress is made.

\paragraph{Random Probing.}

We start strategy discovery with an empty pool of strategies $S = \emptyset$.
A straightforward way to make sure that a new strategy substantially contributes to the current pool $S$
is to always try to solve a problem not yet solved (or \emph{covered}) by any strategy collected so far.
We repeatedly pick such a problem and try to solve it using a \emph{randomly sampled} strategy
out of the totality of all available strategies $\mathbb{S}$. The sampling distribution may be adapted 
to prefer option values that were successful in the past (cf.~Sect.~\ref{sec:sampling_distrib}).
This stage is computationally demanding, but can be massively parallelized.

\paragraph{Strategy Optimization.}

Each newly discovered strategy $s$, solving an as-of-yet uncovered problem $p$, will get 
\emph{optimized} to be as fast as possible at solving $p$. One explores the strategy neighborhood
by iterating over the options (possibly in several rounds), varying option values, and committing to changes that lead to 
a (local) improvement in terms of solution time or, as a tie-breaker, to a default option value where time differences seem negligible. 
We evaluate the impact of this stage in~Sect.~\ref{sec:minimizer}.

\paragraph{Strategy Evaluation.}
\label{sect:StrategyEvaluation}

In the final stage of the discovery process, having obtained an optimized version $s'$ of $s$,
we evaluate $s'$ on all our problems $P$.\todo{Would we do this even if we discovered one strategy $s'$ twice? Note that we assume the determinism of Vampire.}
(This is another computationally expensive, but parallelizable step.)
Thus, we enrich our pool and update
our statistics about covered problems.
Note that every strategy $s'$ we obtain this way
is associated with the problem $p_{s'}$ for which it was originally discovered. We will call this problem 
the \emph{witness problem} of $s'$.



\paragraph{Schedule Construction} can be tried as soon as a sufficiently rich (cf.~Sect.~\ref{sect:discover}) 
pool of strategies is collected. Since we, for every collected strategy, know how it behaves on each problem,
we can pose schedule construction as an optimization task to be solved, e.g., by a (mixed) integer programming (MIP) solver.

In more detail: We seek to allocate time slices $t_s > 0$ to some of the strategies $s\in S$ to cover as many problems as possible while remaining in sum below a given time budget $T$ \cite{DBLP:conf/mkm/HoldenK21,DBLP:conf/paar/Schurr22}.
Alternatively, we may try to cover all the problems known to be solvable in as little total time as possible.\footnote{Strictly speaking,
these only give us a set of strategy-time pairs (as opposed to a sequence).
However, the strategies can be ordered heuristically afterward.}
In this paper, we describe an alternative schedule construction method based on a greedy heuristic,
with a polynomial running time guarantee and other favorable properties (Sect.~\ref{sec:greedy_enters_the_scene}).




%


\subsection{CPU Instructions as a Measure of Time}

We will measure computation time in terms of the number of user instructions executed
(as available on Linux systems through the \texttt{perf} tool).
This is, in our experience, more precise and more stable (on architectures with many cores and many concurrently running processes)
than measuring real 
time.\footnote{For a more thorough motivation, see Appendix A of \cite{EasyChair:7719}.}

In fact, we report \emph{megainstructions (Mi)}, where \SI{1}{Mi} = $2^{20}$ instructions reported by \texttt{perf}.
On contemporary hardware, \SI{2000}{Mi} will typically get used up in a bit less than a second and \SI{256000}{Mi} in around 2 minutes of CPU time.
We also set 
\SI{1}{Mi} as the granularity for the time limits in our schedules.


\section{Strategy Discovery Experiment}
\label{sect:discover}

Following the recipe outlined in Sect.~\ref{sec:spider}, we set out to collect a pool of Vampire (version 4.8) strategies 
covering the first-order form (FOF) fragment of the TPTP library \cite{Sutcliffe2017} version 8.2.0. We focused only 
on proving, so left out all the problems known to be satisfiable, which left us with a set $P$ of \num{7866} problems.
Parallelizing the process where possible, we strived to fully utilize \num{120} cores (AMD EPYC 7513, \SI{3.6}{\giga\hertz}) 
of our server equipped with \SI{500}{\giga\byte} RAM.\todo{We don't mention the specs of the machines we used for schedule construction.}



\begin{figure}[t]
\centering
\includegraphics[scale=0.7]{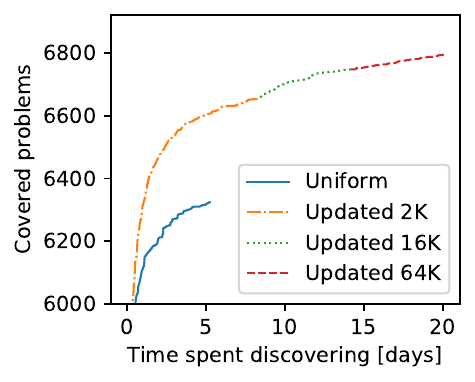}
\includegraphics[scale=0.7]{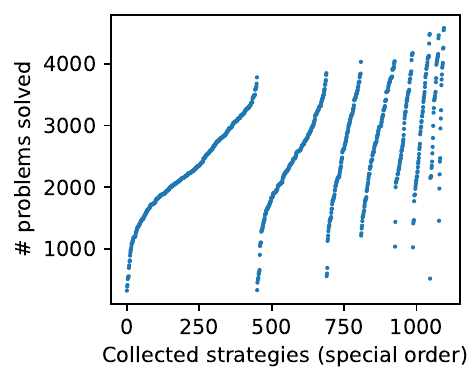}
\caption{Strategy discovery. \emph{Left:} problem coverage growth in time (uniform strategy sampling distribution vs.~an updated one).
\emph{Right:} collected strategies ordered by limit (\num{2000}, \num{4000}, \ldots, \Mi{256000}) and, secondarily, by how many problems can each solve.
}
\label{fig:stamped_coverage}
\end{figure}

We let the process run for a total of \num{20.1} days, in the end covering \num{6796} problems, 
as plotted in Fig.~\ref{fig:stamped_coverage} (left).
The effect of diminishing returns is clearly visible; however, we cannot claim we have exhausted 
all the possibilities. In the last day alone, 8 strategies were added and 9 new problems were solved.


The rest of Fig.~\ref{fig:stamped_coverage} is gradually explained in the following
as we cover some important details regarding the strategy discovery process.

\subsection{Initial Strategy and Varying Instruction Limits}


We seeded the pool of strategies by first evaluating Vampire's default strategy 
for the maximum considered time limit of \SI{256000}{Mi},
solving 4264 problems out of the total 7866.



To save computation time, we did not probe or evaluate all subsequent strategies
for this maximum limit. Instead, to exponentially prefer low limits to high ones,
we made use of the Luby sequence\footnote{\url{https://oeis.org/A182105}} \cite{DBLP:journals/ipl/LubySZ93}
known for its utility in the restart strategies of modern SAT solvers. Our own use of the sequence was as follows.

The lowest limit was initially set to \SI{2000}{Mi} and, multiplying the Luby sequence members
by this number, we got the progression 2000, 2000, 4000, 2000, 2000, 4000, 8000, \ldots as the 
prescribed limits for subsequent probe iterations.
This sequence reaches \SI{256000}{Mi} for the first time in 255 steps. At that point, we stopped following the 
Luby sequence and instead started from the beginning (to avoid eventually reaching limits higher than \SI{256000}{Mi}).

After four such cycles, the lowest, that is \SI{2000}{Mi}, limit probes stopped producing new solutions
(a sampling timeout of \SI{1}{hour} per iteration was imposed). Here, after almost 8.5 days, the ``updated 2K'' plot ends in Fig.~\ref{fig:stamped_coverage} (left). We then increased the lowest limit to \SI{16000}{Mi} and continued in an analogous 
fashion for 155 iterations and 5.7 more days (``updated 16K'') and eventually increased the lowest limit to \SI{64000}{Mi} (``updated 64K'') until the end.


Fig.~\ref{fig:stamped_coverage} (right) is a scatter plot showing the totality of \num{1096} 
strategies that we finally obtained and how they individually perform. The primary order
on the $x$ axis is by the limit and allows us to make a rough comparison of the number of strategies in each limit group 
(\SI{2000}{Mi}, \SI{4000}{Mi}, \ldots, \SI{256000}{Mi}, from left to right).
It is also clear that many strategies (across the limit groups) are, in terms of problem coverage, individually very weak,
yet each at some point contributed to solving a problem considered (at that point) challenging.



\subsection{Problem Sampling}
\label{sec:probsampling}

While the guiding principle of random probing is to constantly aim for solving an as-of-yet unsolved problem,
we modified this criterion slightly to produce a set of strategies better suited for an unbiased
estimation of schedule performance on unseen problems (as detailed in the second half of this paper\todo{Replace by a concrete reference.}).

Namely, in each iteration $i$, we ``forgot'' a random half $P^F_i$ of all problems $P$, considered only
those strategies (discovered so far) whose witness problem lies in the remaining half $P^R_i = P\setminus P^F_i$,
and aimed for solving a random problem in $P^R_i$ not covered by any of these strategies.
This likely slowed the overall growth of coverage, as many problems would need to be covered several times
due to the changing perspective of $P^R_i$. However, we got a (probabilistic) guarantee that any (not too small) subset
$P' \subseteq P$ will contain enough witness problems such that their corresponding strategies will cover $P'$ well.

\subsection{Strategy Sampling}
\label{sec:sampling_distrib}

We sampled a random strategy by independently choosing a random value for each option.
The only exception were dependent options. For example, it does not make sense to configure
the AVATAR architecture (changing options such as \texttt{acc}, which enables congruence closure under AVATAR) if the main AVATAR option (\texttt{av}) is set to \texttt{off}.
Such complications can be easily avoided by following, during the sampling, 
a topological order that respects the option dependencies. (For example, we sample \texttt{acc} only after the value \texttt{on} has been chosen for \texttt{av}.)

Even under the assumption of option independence, the mean time in which a random strategy solves
a new problem can be strongly influenced by the value distributions for each option.
This is because some option values are rarely useful and may even substantially reduce the prover performance,
for example, if they lead to a highly incomplete strategy.\footnote{An extreme example is turning off \emph{binary resolution},
the main inference for non-equational reasoning. This can still be useful, for instance 
when replaced by \emph{unit resulting resolution} \cite{DBLP:conf/cav/KovacsV13}, but our sampling needs to discover this by chance.}
Nevertheless, not to preclude the possibility of discovering arbitrarily wild strategies, 
we initially sampled every option uniformly where possible.\footnote{Exceptions were: 1. The ratios: 
e.g., for \texttt{age\_weight\_ratio} we sampled uniformly its binary logarithm (in the range $-10$ and $4$)
and turned this into a ratio afterward (thus getting values between $1:1024$ and $16:1$); 2. Unbounded integers
(an example being the \emph{naming threshold} \cite{DBLP:conf/gcai/Reger0V16}), for which we used a geometric distribution instead.}


Once we collected enough strategies,\footnote{This was done in an earlier version of the main experiment.} we updated the frequencies for sampling finite-domain options
(which make up the majority of all options) by counting how many times each value occurred in a strategy that, at the moment
of its discovery, solved a previously unsolved problem. (This was done before a strategy got optimized. Otherwise
the frequencies would be skewed toward the default, especially for option values that rarely help
but almost never hurt.) 

The effect of using an updated sampling distribution for strategy discovery can be seen in Fig.~\ref{fig:stamped_coverage} (left).
We ran two independent versions of the discovery process, one with the uniform distribution and one with the updated distribution.
We abandoned the uniform one after approximately 5 days, by which time it had covered \num{6324} problems
compared to \num{6607} covered with the help of the updated distribution at the same mark.
%
We can see that the rate at which we were able to solve new problems became substantially higher with the updated distribution.


\subsection{Impact of Strategy Optimization}
\label{sec:minimizer}

Once random probing finds a new strategy $s$ that solves a new problem $p$,
the task of optimization (recall Sect.~\ref{sec:spider}) is to search the option-value neighborhood of $s$ for a strategy $s'$
that solves $p$ in as few instructions as possible and preferably uses default option values (where this does not compromise performance on $p$).

\begin{figure}[t]
\centering
\includegraphics[scale=0.7]{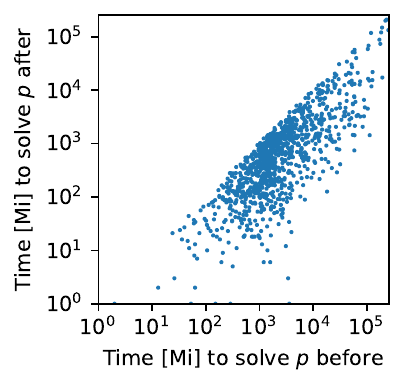}
\includegraphics[scale=0.7]{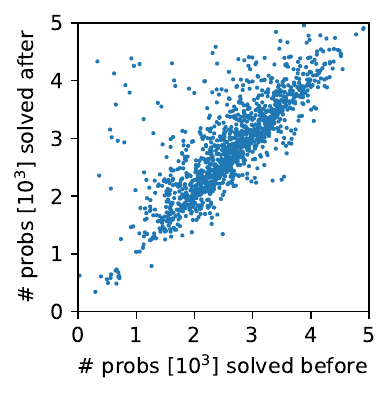}
\caption{Strategy optimization scatter plots. \emph{Left:} time needed to solve strategy's witness problem (a log--log plot).
\emph{Right:} the total number of problems (in thousands) solved.}
\label{fig:optimize}
\end{figure}



The impact of optimization is demonstrated in Fig.~\ref{fig:optimize}. On the left, we can see that, almost invariably,
optimization substantially improves the performance of the discovered strategy on its witness problem $p$. The geometric mean 
of the improvement ratio we observed was \num{4.2} (and the median \num{3.2}). The right scatter plot shows the overall performance of each strategy.\footnote{In a previous version of the main experiment,
we evaluated each strategy both before and after optimization, which gave rise to this plot.} Here, the observed 
improvement 
is $\times$1.09 on average (median \num{1.03}), and the improvement is solely an effect of setting option values 
to default where possible (without this feature, we would get a geometric mean of the improvement 0.84 and median 0.91).
In this sense, the tendency to pick default values regularizes the strategies, making them more powerful also on problems other than their witness problem.

\subsection{Parsing Does Not Count}
\label{sec:pdnc}

When collecting the performance data about the strategies, we decided to ignore the time it takes Vampire to
parse the input problem. This was also reflected in the instruction limiting, so that running Vampire
with a limit of, e.g., \SI{2000}{Mi} would allow a problem to be solved if it takes at most \SI{2000}{Mi}
on top of what is necessary to parse the problem.

The main reason for this decision is that Vampire, in its strategy scheduling mode, starts dispatching 
strategies only after having parsed the problem, which is done only once. Thus, from the perspective 
of individual strategies, parsing time is a form of a sunk cost, something that has already been paid.

Although more complex approaches to taking parse time into account when optimizing
schedules are possible, we in this work simply pretend that problem parsing always costs 0 instructions. This should be taken
into account when interpreting our simulated performance results reported next (in Sect.~\ref{sec:casc}, but also in Sect.~\ref{sec:experimental}).\iftoggle{PREPRINT}{\footnote{Appendix~\ref{sec:empirical} shows that these do not depart too much from an empirical evaluation.}}






\section{One Schedule to Cover Them All} 
\label{sec:casc}

\begin{figure}[t]
\centering
\includegraphics[scale=0.7]{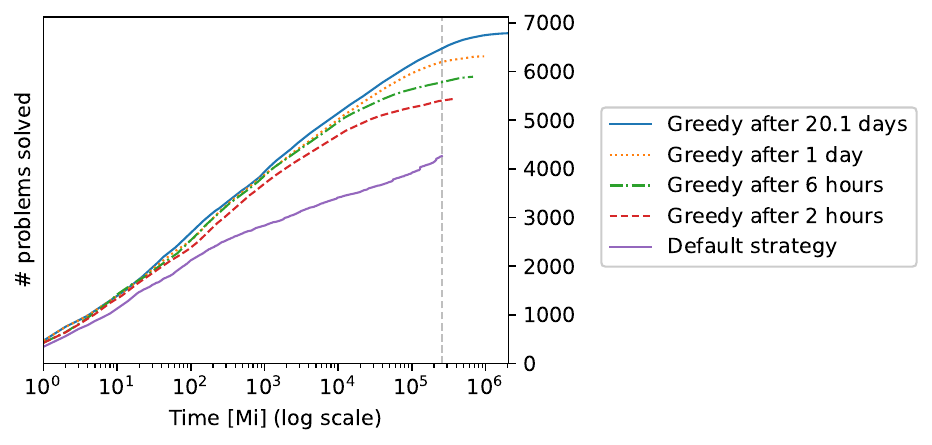}
\caption{Cumulative performance 
of several greedy schedules,
each using a subset of the discovered strategies as gathered in time,
compared with Vampire's default strategy}
\label{fig:vbs}
\end{figure}


Having collected our strategies, let us pretend that we already know how to construct a schedule
(to be detailed in Sect.~\ref{sec:greedy}) and use this ability to answer some imminent questions,
most notably: How much can we now benefit?

Fig.~\ref{fig:vbs} plots the cumulative performance (a.k.a.~``cactus plot'') of schedules we could build after \SI{2}{hours},
\SI{6}{hours}, \SI{1}{day}, and full \SI{20.1}{days} of the strategy discovery. 
The dashed vertical line denotes the time limit of \SI{256000}{Mi}, which roughly corresponds to a 2-minute prover run.
For reference, we also plot the behavior of Vampire's default strategy. We can see that already after two hours
of strategy discovery, we could construct a schedule improving 
on the default strategy by \SI{26}{\percent} (from \num{4264} to \num{5403} problems solved).
Although the added value per hour spent searching gradually drops,
the \SI{20.1}{days} schedule is still \SI{4}{\percent} better
than the \SI{1}{day} one (improving from 6197 to 6449 
at \SI{256000}{Mi}).\iftoggle{PREPRINT}{\footnote{
A performance comparison of a general schedule, such as the one shown here, with just greedily stacked equally sized strategy slices
has been moved to Appendix~\ref{sec:vbs}.}}

The plot's $x$-axis ends at $8\cdot$\SI{256000}{Mi}, which roughly corresponds to the time limit used 
by the most recent CASC competitions \cite{Sut16} in the FOF division (i.e., 2 minutes on 8 cores\todo{How does Vampire utilize the cores when the schedule is just one sequence?}).
The strongest schedule shown in the figure manages to solve \num{6789} problems (of the \num{6796} covered in total) at that mark.\footnote{It is possible to solve all 6796 covered problems with a schedule that spans \Mi{2582228}. This is the optimum length -- no shorter schedule solving all covered problems exists.}
%
We remark that this schedule, in the end, employs only \num{577} of the \num{1096} available strategies,
which points towards a noticeable redundancy in the strategy discovery process.\todo{How many strategies does the schedule covering all problems in minimum time use? It may be a better proof of redundancy.}

One way to fit all the solvable problems below the CASC budget 
would be to use a standard trick and split the totality of problems $P$ into two or more easy-to-define syntactic classes (e.g., 
Horn problems, problems with equality, large problems, etc.) and construct
dedicated schedules for each class in isolation. The prover can then be dispatched
to run an appropriate schedule once the input problem features are read.
We do not explore this option here.
Intuitively, by splitting $P$ into smaller subsets, the risk of overfitting 
to just the problems
for which the strategies were discovered increases, and we mainly want to explore here the opposite, 
the ability of a schedule to generalize to unseen problems.




\section{Greedy Schedule Construction}
\label{sec:greedy_enters_the_scene}
\label{sec:greedy}

Having collected a set of strategies $S$ and evaluated each on the problems in $P$, let us 
by $\SolveTimeP{s}{p} : S \times P \to \Nat \cup \{\infty\}$ denote the \emph{evaluation matrix}\todo{FB: I find the mixing of matrix and function notation confusing.},
which records the obtained solutions times (and uses $\infty$ to signify a failure 
to solve a problem within the evaluation time limit used). Given a time budget $T$,
the \emph{schedule construction problem} (SCP) is the task of assigning a time limit $t_s \in \Nat$ to every strategy $s \in S$, such that the number of covered problems
\[\left\lvert\bigcup_{s \in S } \{p \in P \,|\, \SolveTimeP{s}{p} \leq t_s \}\right\rvert,\]
subject to the constraint $\sum_{s \in S} t_s \leq T$, is maximized.
\label{sec:schedule_construction_problem}

To obtain a schedule as a sequence (as defined in Sect.~\ref{sec:prelim}),
we would need to order the strategies having $t_s > 0$. This can, in practice, be done 
in various ways, but since the order does not influence the predicted 
performance of the schedule under the budget $T$, we keep it here unspecified
(and refer to the mere time assignment $t_s$ as a \emph{pre-schedule} where the distinction matters).

\begin{algorithm}[t]
\caption{Greedy schedule construction (base version)}
\label{alg:greedy}
\begin{algorithmic}[1]
\Require Problems $P$, strategies $S$, solution times $\SolveTimeP{s}{p}$, time budget $T$
\Ensure Pre-schedule $t_s : S \to \Nat$
\State $t_s \gets \mathbf{0}$  \Comment{Start with zeros everywhere}
\State $T' \gets T, P' \gets P$ \Comment{Remaining budget, remaining problems}
\While{the maximum just below is positive}
\State $s, t \gets \argmax_{s\in S, t \in \Nat, 0 < (t - t_s) \leq T' } |\{p \in P' \,|\, \SolveTimeP{s}{p} \leq t \}| / (t - t_s)$
\label{line:criterion}
\State $T' \gets T' - (t - t_s)$ \Comment{Update the remaining budget}
\State $t_s \gets t$ \Comment{Extend the pre-schedule}
\State $P' \gets \{p \in P' \,|\, \SolveTimeP{s}{p} > t\}$ \Comment{Remove covered problems from $P'$}
\EndWhile
\end{algorithmic}
\end{algorithm}

Although it is straightforward to encode SCP as a mixed integer program 
and attempt to solve it exactly (though it is an NP-hard problem), an adaptation 
of a greedy heuristic from a closely related (budgeted) maximum coverage problem \cite{DBLP:journals/mor/Chvatal79,DBLP:journals/ipl/KhullerMN99} works surprisingly well in practice and runs in time polynomial in the size of $\SolveTimeP{s}{p}$.
The key idea is to greedily maximize the number of newly covered problems \emph{divided} by the amount of time this additionally requires.

\cref{alg:greedy} shows the corresponding pseudocode. It starts from an empty schedule $t_s$
and iteratively extends it in a greedy fashion. The key criterion appears on line~\ref{line:criterion}.
Note that this line corresponds to an iteration over all available strategies $S$ and, for each strategy $s$, all meaningful time limits
(which are only those where a new problem gets solved by $s$, so their number is bounded by $|P|$).

\cref{alg:greedy} departs from the obvious adaptation of the above-mentioned greedy algorithm for the set covering problem \cite{DBLP:journals/mor/Chvatal79} in that we allow extending a slice of a strategy $s$ that is already included in the schedule
(that is, has $t_s > 0$) and ``charge the extension'' only for the additional time it claims (i.e., $t - t_s$). This \emph{slice extension}
trick turns out to be important for good performance.\iftoggle{BUDGETS}{\footnote{The degree of importance of slice extension can be observed in \cref{fig:ShortBudgets} in \cref{sec:ExperimentsExtra}.}}{\todo{Briefly discuss how important it is.}}

\subsection{Do We Need a Budget?}

A budget-less version of \cref{alg:greedy} is easy to obtain (imagine $T$ being very large).
When running on real-world $\SolveTimeP{s}{p}$ (from evaluated Vampire strategies), we noticed that
the length of a typical extension $(t - t_s)$ tends to be small relative to the current used-up time $\sum_{s\in S}t_s$ and that
the presence of a budget starts affecting the result only when the used-up time comes close to the budget.

As a consequence, if we run a budget-less version and, after each iteration, record the 
pair $(\sum_{s\in S}t_s,|P\setminus P'|)$, we get a good estimate (in a single run) of 
how the algorithm would perform for a whole (densely inhabited) sequence of relevant budgets.
This is how the plot in Fig.~\ref{fig:vbs} was obtained. Note that this would be
prohibitively expensive to do when trying to solve the SCP optimally.

We can also use this observation in an actual prover. 
If we record and store a journal of the budget-less run,
remembering which strategy got extended in each iteration and by how much,
we can, given a concrete budget $T$, quickly replay the journal just to the point 
before filling up $T$, and thus get a good schedule for the budget $T$ without having to optimize specifically for $T$.


\section{Regularization in Schedule Construction}
\label{sec:regul}

To estimate the future performance of a constructed schedule on previously unseen problems,
we adopt the standard methodology used in statistics, randomly split our problem set $P$ 
into a train set $P_\mli{train}$ and a test set $P_\mli{test}$, construct a schedule for the first, and evaluate it on the second. 

To reduce the variance in the estimate, we use many such random splits 
and average the results. In the experiments reported in the following, we actually compute
an average over several rounds of 5-fold cross-validation \cite{DBLP:books/lib/HastieTF09}.
This means that the size of $P_\mli{train}$ 
is always \pc{80} and the size $P_\mli{test}$ \pc{20} of our problem set $P$. However,
we \emph{re-scale} the reported train and test performance back to the size of the whole problem set $P$ 
to express them in units that are immediately comparable.
We note that the reported performance is obtained though simulation, i.e.,
it is only based on the evaluation matrix $\SolveTimeP{s}{p}$.

\label{sec:train_strategies_hygiene}

\paragraph{Training Strategy Sets.}
We retroactively simulate the effect of discovering strategies only for current training problems $P_\mli{train}$.
Given our collected pool of strategies $S$, we obtain the training strategy set $S_{\mli{train}}$ 
by excluding those strategies from $S$ 
whose witness problem lies outside $P_{\mli{train}}$ (cf.~Sect.~\ref{sec:probsampling}).
When a schedule is optimized on the problem set $P_\mli{train}$,
the training data consists of the results of the evaluations of strategies $S_{\mli{train}}$ on problems $P_{\mli{train}}$.




%

\subsection{Regularization Methods}

We propose several modifications of greedy schedule construction (\cref{alg:greedy})
with the aim of improving its performance on unseen problems (the test set performance)
while possibly sacrificing some of its training performance.


With the base version, we observed that it could often solve more test problems
by assigning more time to strategies introduced into the schedule in early iterations,
at the expense of strategies added later
(the latter presumably covering just a few expensive training problems and being over-specialized to them).
Most of the modifications described next
assign more time to strategies added during early iterations,
each according to a different heuristic.

\begin{description}
\item[Slack.]
The most straightforward regularization we explored
extends each non-zero strategy time limit $t_s$ in the schedule
by multiplying it by the multiplicative slack $w \geq 1$
and adding the additive slack $b \in \{0, 1, \ldots\}$.
For each $t_s > 0$, the new limit $t_s'$ is therefore $t_s \cdot w + b$.
To avoid overshooting the budget,
we keep track of the total length of the extended schedule during the construction
(implementation details are slightly more complicated but not immediately important).
The parameters $w$ and $b$ control the degree of regularization, and with $w=1$ and $b=0$, we get the base algorithm.
\todo[inline]{Future research: Describe this as a post-processing step and apply to Gurobi schedules.}


\item[Temporal Reward Adjustment.]
\todo{Consider renaming. Other ideas for a name: reward delaying, delayed gratification, polynomial discounting, reward exponentiation.}
\label{sec:temporal_reward_adjustment}
In each iteration of the base greedy algorithm,
we select a combination of strategy $s$ and time limit $t$
that maximizes the number of newly solved problems $n$\todo{Use $n$ in the base algorithm definition.} per time $t$.
Intuitively, the relative degree to which these two quantities influence the selection is arbitrary.
To allow stressing $n$ more or less with respect to $t$,
we exponentiate $n$ by a regularization parameter $\alpha \geq 0$,
so the decision criterion becomes $\frac{n^\alpha}{t}$.

For small values of $\alpha$,
the algorithm values the time more and becomes eager to solve problems early.
For large values of $\alpha$, on the other hand,
the algorithm values the problems more and prefers longer slices that cover more problems.
For example, for $\alpha = 1.5$,
the algorithm prefers solving 2 problems in \Mi{5000} 
to solving 1 problem in \Mi{2000}.
Compare this to $\alpha = 1$ (the base algorithm),
which would rank these slices the other way around.


\item[Diminishing Problem Rewards.]
By covering a training problem with more than one strategy, we cover it robustly:
When a similar testing problem is solved by only one of these strategies,
the schedule still manages to solve it.
However, the base greedy algorithm does not strive to cover any problem more than once:
as soon as a problem is covered by one strategy,
this problem stops participating in the scheduling criterion.
This is the case even when covering the problem again would cost relatively little time.

Regularization by diminishing problem rewards covers problems robustly
by rewarding strategy $s$
not only by the number of \emph{new} problems it covers
but also by the problems covered by $s$ that are already covered by the schedule.
This is achieved by modifying the slice selection criterion.
Instead of maximizing the number of new problems solved per time,
we maximize the total reward per time,
which is defined as follows:
Each problem contributes the reward $\beta^k$,
where $k$ is the number of times the schedule has covered the problem
and $\beta$ is a regularization parameter ($0 \leq \beta \leq 1$).
We define $0^0 = 1$ so that $\beta = 0$ preserves the original behavior of the base algorithm.

For example, for $\beta = 0.1$,
each problem contributes the reward $1$ the first time it is covered,
$0.1$ the second time,
$0.01$ the third time, etc.
Informally, the algorithm values
covering a problem the second time in time $t$ as much as
covering a new problem in time $10\cdot t$.



\end{description}
These modifications are independent and can be arbitrarily combined.

\subsection{Experimental Results}
\label{sec:experimental}

We evaluated the behavior of the previously proposed techniques using three
time budgets: \Mi{16000} ($\approx$\SI{8}{\second}), \Mi{64000}
($\approx$\SI{32}{\second}), and \Mi{256000}
($\approx$\SI{2}{\minute}).\iftoggle{PREPRINT}{\footnote{See Appendix \ref{sec:stamped} and \ref{sec:cherries} for 
some additional perspectives on how the test performance of the base greedy algorithm develops 
when varying the budget.}}

\paragraph{Optimal Schedule Constructor.}
In the existing approaches to the construction of strategy schedules \cite{DBLP:conf/mkm/HoldenK21,DBLP:conf/paar/Schurr22},
it is common to encode the SCP (see \cref{sec:schedule_construction_problem}) as a mixed-integer program
and use a MIP solver to find an exact solution.
We implemented such an optimal schedule construction (OSC) by encoding the problem\footnote{\iftoggle{PREPRINT}{We used an encoding described in \cref{sec:encodeSCP}.}{We used a straightforward encoding similar to the encoding described by Holden and Korovin \cite{DBLP:conf/mkm/HoldenK21}.}} in Gurobi~\cite{gurobi} (ver.~10.0.3)
and compared  OSC to the base greedy schedule construction
(\cref{alg:greedy}) on 10 random $80:20$ splits.

For the budget of \Mi{256000}, it takes Gurobi over \SI{16}{\hour}
to find an optimal schedule, whereas the greedy algorithm finds a schedule in less
than a minute. The optimal schedule solves, on average, $45.0$ (resp.~$8.5$) more
problems than the greedy schedule on $P_\mli{train}$ (resp.~on
$P_\mli{test}$) when re-scaled to $|P|$.\todo{How many problems in absolute numbers?\newline KC: $6536.7$ and $6057.4$ I agree it should be there, but I'm reluctant to do that due to incompatible splits and hence numbers with~\cref{tab:schedulers}.}
%
For the \Mi{16000} and \Mi{64000} budgets, Gurobi does not solve the
optimal schedule within a reasonable time limit.
For example, after
\SI{24}{\hour}, the relative gaps between the lower and upper objective
bound are \SI{5.38}{\percent} and
\SI{1.43}{\percent}, respectively. This makes the OSC
impractical to use as a baseline for our
regularization experiments.\footnote{A better encoding and solver settings may
  improve on this. However, we suspect the problem to be hard; we
  tried some modifications with similar (or worse) results.}

\begin{figure}[t]
    \centering
    \includegraphics[width=\linewidth]{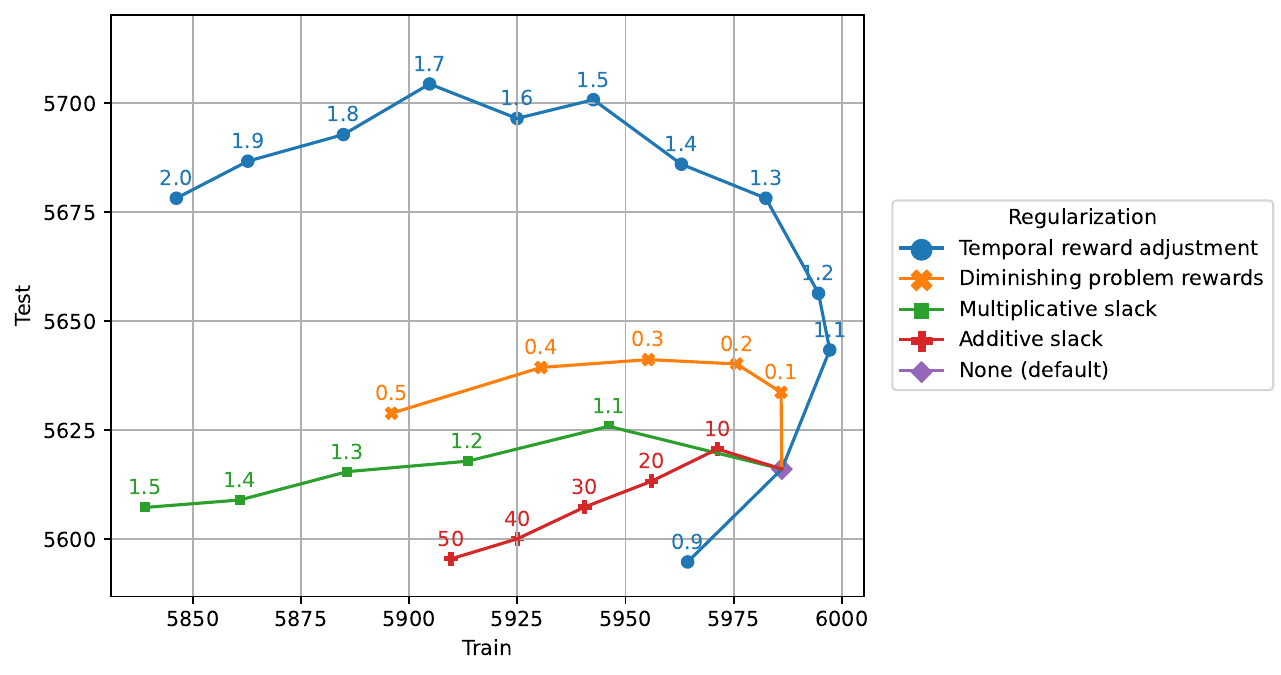}
    \caption{Train and test performance of various regularizations of the greedy schedule construction algorithm for the budget \Mi{64000}.
    Performance is the mean number of problems solved out of \num{7866} across 50 splits.
    The label of each point denotes the value of the respective regularization parameter.
    }
    \label{fig:snakes}
\end{figure}

\paragraph{Regularization of the Greedy Algorithm.}
\label{sec:experiment_regularization}
To estimate the performance of the proposed regularization methods,
we evaluated each variant on 50 random splits (10 times 5-fold cross-validation).
We assessed the algorithm's response to each regularization parameter in isolation.
For each parameter, we evaluated regularly spaced values from a promising interval covering the default value ($b=0$, $w=1$, $\alpha=1$, $\beta=0$).
\Cref{fig:snakes} demonstrates the effect of these variations on the train and test performance for the budget \Mi{64000}.\footnote{This budget seems to be the most practically relevant (e.g.,~for the application in interactive theorem provers).\iftoggle{BUDGETS}{ The other two budgets are detailed in \cref{sec:ExperimentsExtra}.}{}}

Temporal reward adjustment was the most powerful of the regularizations,
improving test performance for all the evaluated values of $\alpha$ between 1.1 and 2.0.
Surprisingly, the values 1.1 and 1.2 also improved the train performance,
suggesting that the default greedy algorithm is too time-aggressive on our dataset.\todo{Speculate why this regularization is so good.}
\todo[inline]{Analyze this regularization in more depth.}
\todo[inline]{For each interesting configuration, plot the \enquote{anytime budget} cactus curve and compare. This way, we could show, for example, that $\alpha=1.7$ is better than $\alpha=0$ for all budgets.}

\Cref{tab:schedulers} compares the performance of notable configurations of the greedy algorithm.
Specifically, we include evaluations of
the base greedy algorithm and
the best of the evaluated parameter values for each of the regularizations.
The table also illustrates the effect of regularizations on the computational time of the greedy schedule construction:
$\beta > 0$ slows the procedure down and $\alpha > 1$ speeds it up.

\newcommand{\perf}{\num[round-mode=places,round-precision=0]}
\newcommand{\ttf}{\num[round-mode=places,round-precision=0]}

\begin{table}[t]
    \centering
    \caption{Comparison of regularizations of the greedy schedule construction algorithm for the budget \Mi{64000}.
    Performance is the mean number of problems solved out of \num{7866} across 50 splits.
    Time to fit is the mean time to construct a schedule in seconds.}
    \begin{tabular}{@{}c|rr|r@{}}
        \toprule
        \multirow{2}{*}{Regularization} & \multicolumn{2}{l}{Performance} & \multirow{2}{*}{Time to fit [s]} \\
        & Test & Train & \\
        \midrule


$\alpha = 1.7$ & \perf{5704.399593} & \perf{5904.724703} &  \ttf{4.325128} \\  
$\beta = 0.3$  & \perf{5641.199691} & \perf{5955.299844} & \ttf{66.538374} \\  
$w = 1.1$      & \perf{5625.900160} & \perf{5946.099763} & \ttf{19.332677} \\  
$b = 10$       & \perf{5620.700770} & \perf{5971.249778} & \ttf{19.607397} \\  
None (default) & \perf{5616.100439} & \perf{5986.124790} & \ttf{19.885694} \\  

        \bottomrule
    \end{tabular}
    \label{tab:schedulers}
\end{table}
\todo[inline]{Consider including the variance of the performance scores.}

In a subsequent experiment,
we searched for a strong
combination of regularizations by
local search from the strongest single-parameter regularization ($\alpha = 1.7$).\todo{Consider discussing joint optimization using Nevergrad.}
This yielded a negligible improvement over $\alpha = 1.7$:
The best observed test performance was 5707 ($\alpha = 1.7$ and $b = 30$), compared to 5704 of $\alpha = 1.7$.

Finally, we briefly explored
the interactions between the budget and the optimal values of the regularization parameters.
For each of the three budgets of interest and each of the regularization parameters,
we identified the best parameter value from the evaluation grid.
\Cref{tab:budgets} shows that the best configurations of all the regularizations except multiplicative slack vary across budgets.\iftoggle{BUDGETS}{\footnote{See a more detailed comparison in \cref{sec:ExperimentsExtra}.}}

\newcolumntype{C}{>{\raggedleft\arraybackslash}p{1cm}}

\begin{table}[t]
    \centering
    \caption{Best observed values of regularization parameters for various budgets}
    \begin{tabular}{@{}r|CCCC@{}}
        \toprule
        \multirow{2}{*}{Budget [Mi]} & \multicolumn{4}{c}{Best parameter value} \\
        & $b$ & $w$ & $\alpha$ & $\beta$ \\
        \midrule
         16000 &  0 & 1.1 & 1.3 & 0.2 \\
         64000 & 10 & 1.1 & 1.7 & 0.3 \\
        256000 & 20 & 1.1 & 1.4 & 0.2 \\
        \bottomrule
    \end{tabular}
    \label{tab:budgets}
\end{table}

\todo[inline]{The optimum combination of regularizers seems to be budget-dependent. Include a plot that shows this -- how the relative performance of schedulers varies across budgets.}

\section{Related Work}



Outside the realm of theorem proving, strategy discovery belongs 
to the topic of \emph{algorithm configuration} \cite{DBLP:journals/jair/SchedeBTWBHT22},
where the task is to look for a strong configuration of a parameterized algorithm automatically.
Prominent general-purpose algorithm configuration procedures include ParamILS \cite{DBLP:journals/jair/HutterHLS09} and SMAC \cite{DBLP:journals/jmlr/LindauerEFBDBRS22}. 





To gather a portfolio of complementary configurations, 
Hydra \cite{DBLP:conf/aaai/XuHL10} searches for them in rounds, 
trying to maximize the marginal contribution against all the configurations identified previously.
Cedalion \cite{DBLP:conf/aaai/SeippSHH15} is interesting in that it maximizes such contribution \emph{per unit time},
similarly to our heuristic for greedy schedule construction. 
Both have in common that they, a priori, consider all the input problems in their criterion.
BliStr and related approaches \cite{DBLP:conf/gcai/Urban15,DBLP:conf/cpp/JakubuvU17,DBLP:conf/gcai/JakubuvSU17,DBLP:conf/mkm/HoldenK21,DBLP:conf/mkm/HulaJJK22}, on the other hand, combine strategy improvement with problem clustering 
to breed strategies that  are ``local experts'' on similar problems. Spider \cite{Spider} is even more radical in
this direction and optimizes each strategy on a single problem.\footnote{Although the preference for default option values
as a secondary criterion, at the same time, helps to push for good general performance (see Sect.~\ref{sec:minimizer}).}



Once a portfolio of strategies is known, it may be used in one of several ways to solve a new input problem:
execute all strategies in parallel \cite{DBLP:conf/aaai/XuHL10},
select a single strategy \cite{DBLP:conf/mkm/HulaJJK22},
select one of pre-computed schedules \cite{DBLP:conf/mkm/HoldenK21},
construct a custom strategy schedule \cite{DBLP:conf/cade/ManglaHP22},
schedule strategies dynamically \cite{DBLP:journals/jar/KuhlweinU15}, or
use a pre-computed static schedule \cite{DBLP:conf/cpp/JakubuvU17,DBLP:conf/paar/Schurr22}. 
The latter is the approach we explored in this work.

A popular approach to construct a static schedule (besides solving SCP optimally \cite{DBLP:conf/mkm/HoldenK21,DBLP:conf/paar/Schurr22})
is to greedily stack uniformly-timed slices \cite{DBLP:conf/cpp/JakubuvU17}.\footnote{However,
uniformly-timed slices only get close to the performance of our greedy schedule at a small region
depending on the slice time limit used\iftoggle{PREPRINT}{ (see Appendix \ref{sec:vbs})}.} 
Regularization in this context is discussed by Jakubuv et al.~\cite{DBLP:conf/itp/JakubuvCGKOP00U23}. 
Finally, a different greedy approach to schedule construction
was already proposed in \emph{p-SETHEO} \cite{DBLP:conf/flairs/WolfL98}.\iftoggle{PREPRINT}{\footnote{\Cref{sec:psetheo} compares this approach
to ours in some detail.}}{\todo{Compare p-SETHEO to our approach briefly.}}

\section{Conclusion}
\label{sec:conclusion}

In this work, we conducted an independent evaluation of Spider-style \cite{Spider} strategy discovery
and schedule creation. Focusing on the FOF fragment of the TPTP library, we collected over a thousand
Vampire proving strategies, each a priori optimized to perform well on a single problem. 
Using these strategies, it is easy to construct a single monolithic schedule 
which covers most of the problems known to be solvable 
within the budget used by the CASC competition. This suggests that for CASC
not to be mainly a competition in memorization, using a substantial set 
of previously unseen problems each year is essential.

To construct strong schedules using the discovered strategies,
we proposed a greedy schedule construction procedure, which can compete with optimal approaches.
For a time budget of approximately 2 minutes,
the greedy algorithm takes less than a minute to produce a schedule
that solves more than \pc{99} as many problems as an optimal schedule\todo{This data point is not included anywhere else in the paper.}, which takes more than 16 hours to generate.
For shorter time budgets, optimal schedule construction is no longer feasible,
while greedy construction still produces relatively strong schedules.

This surprising strength of the greedy scheduler can be further reinforced by various regularization mechanisms,
which constitute the main contribution of this work.
An appropriately chosen regularization allows us to outperform the optimal schedule 
on unseen problems. 
%
Finally, the runtime speed and simplicity of the greedy schedule construction algorithm and the regularization techniques make them attractive for reuse and further experimentation.


\begin{credits}
\subsubsection{\ackname}
\Acknowledgments
%
\end{credits}

\appendix

\iftoggle{BUDGETS}{
\section{Experimental Results on Various Budgets}
\label{sec:ExperimentsExtra}

In addition to the budget of \Mi{64000} (approx.~\SI{32}{\second}), which we discussed in \cref{sec:experimental},
we evaluated the schedule construction algorithms on the budgets of \Mi{16000} (approx.~\SI{8}{\second}) and \Mi{256000} (approx.~\SI{2}{\minute}).
\Cref{fig:ShortBudgets} shows the results of these evaluations.
It shows namely that temporal reward adjustment is the most powerful of the regularizations under consideration for all of these budgets,
and that the optimal values of most of the regularization parameters vary across budgets.

To demonstrate the effect of the slice extension trick described in \cref{sec:greedy},
we also include two weaker versions of the base greedy algorithm:
one without slice extension
and one with the slice extension restricted to the most recent slice (\enquote{conservative slice extension}).
Both of these modifications allow including any single strategy in the schedule more than once,
which is implemented in a straightforward fashion.

\begin{figure}
	\centering
	\includegraphics[width=\linewidth]{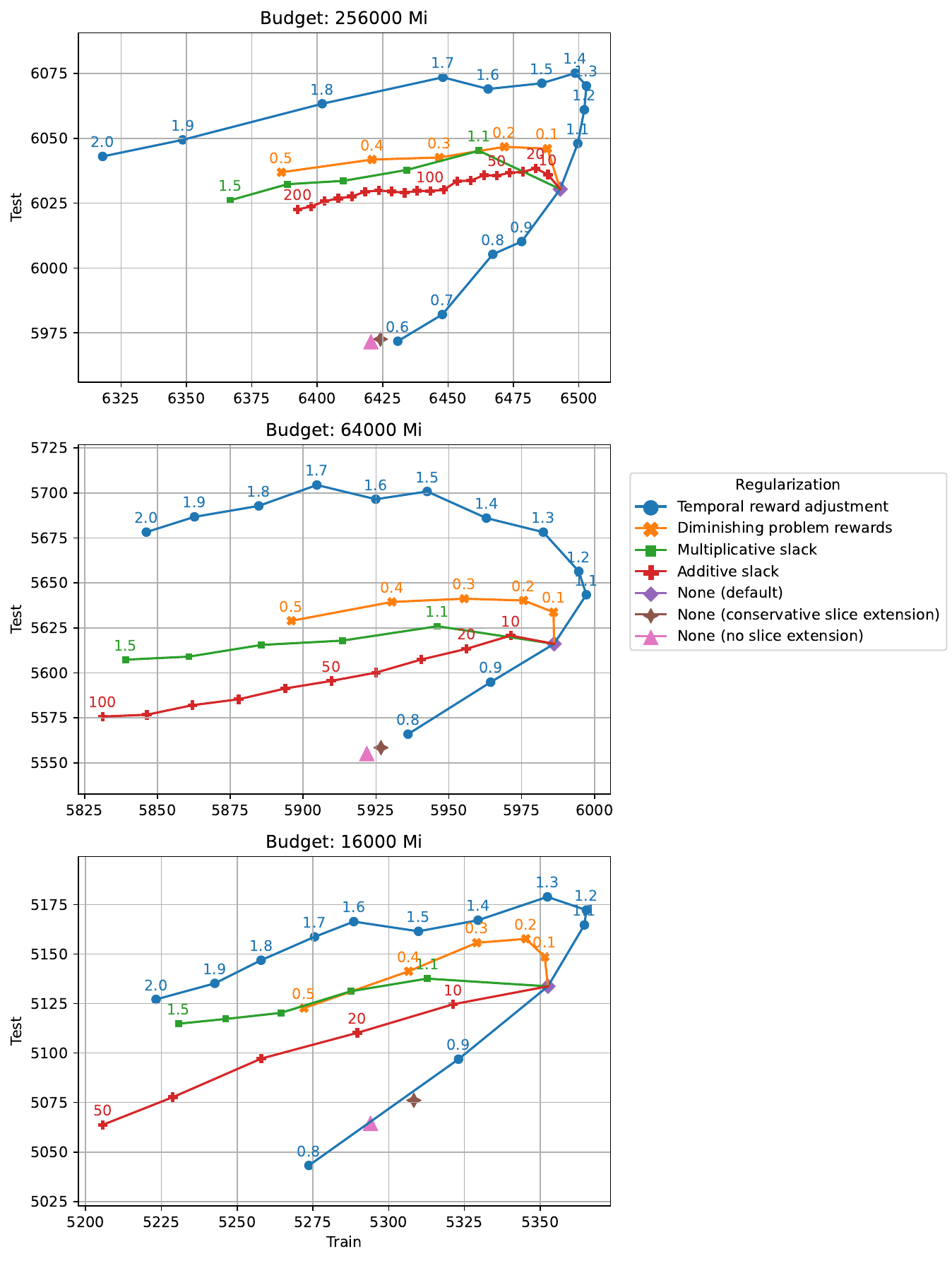}
	\caption{Train and test performance of various regularizations of the greedy schedule construction algorithm for the budgets \Mi{256000} (\emph{top}), \emph{\Mi{64000}} (\emph{middle}), and \Mi{16000} (\emph{bottom}).
		Performance is mean number of problems solved out of 7866 across 50 splits.
		The label of each point denotes the value of the respective regularization parameter.}
	\label{fig:ShortBudgets}
\end{figure}

}

\iftoggle{PREPRINT}{



\section{Evaluating a ``CASC'' Schedule Empirically}
\label{sec:empirical}

To get an idea on how a schedule, built using the real-life data collected 
during the strategy discovery experiment (as described in Sect.~\ref{sect:discover}),
would behave when executed from within Vampire, we proceeded as follows:
\begin{enumerate}
	\item We applied the base greedy algorithm under a budget of \SI{2048000}{Mi}.
	\item We sorted the obtained slices greedily, in the decreasing order of ``number of problems additionally covered divided by slice's time limit.''
	\item Before outputting the slices, we extended each slice by \SI{10}{Mi}. 
	\label{item:extend}
\end{enumerate}
The obtained schedule uses \num{553} strategies, should solve \num{6789} problems,
and requires \SI{2028647}{Mi} to execute. (The budget of \SI{2048000}{Mi} did not get fully used up during construction, so that the additional $10\cdot 553$ of added slack still fit in.)


We then evaluated this schedule from Vampire, using---in parallel---60 out of 72 
(hyperthreaded) cores of our server with 
Intel\textregistered Xeon\textregistered Gold 6140 CPU @ \SI{2.3}{\giga\hertz} CPUs,
under a time limit of \SI{960}{\second} per problem.





Vampire in the end solved \num{6612} problems. 
Out of these, \num{424} were solved by a strategy which was added to the schedule to solve them,
\num{6165} were solved by an earlier strategy in the schedule, 
\num{22} by a later strategy, and there was one problem solved
that the schedule was not meant to cover.

\num{1076} were left unsolved ``as expected'', but \num{178} problems ``went missing''. 
Out of these missing ones, however, 
only on \num{13} Vampire did not manage to fully execute all the slices 
before the time limit and only on one did not finish the slice which was meant to solve the problem. 
This last mentioned was \texttt{CSR111+6}, one of the most expensive problems in
our set to parse (taking approximately \SI{158300}{Mi} to parse).
So here we found a case of ignoring parsing times adversely influencing predictions about schedule execution, but not an explanation for majority of the missing problems.

The explanation rather lies with an occasional variance in the instruction measurements,
which the schedule likes to (ab)use when the measurements look exceptionally good:
Out of the \num{178} missing problems, \num{121} were meant to be solved \SI{10}{Mi} before 
their extended slice's end, i.e., exactly at the end of their original 
slice before it got extended (in step \ref{item:extend}. above). In other words,
these problems were ``sitting on the edge'' to begin with. When manually re-measuring
the solution times for these problems and their strategies, 
we found that, in majority of the cases, the typical value to be higher than what $E^s_p$
predicted, often too high for the added slack of \SI{10}{Mi} to save it.

To mitigate or even completely eliminate this effect for an actual competition,
we recommend re-evalauting the strategies from a constructed schedule
on problems which each respective strategy is aiming to solve (note that this is 
a tiny fraction compared to evaluating all the strategies on all the problems)
and later using an averaged value. Upon subsequent schedule construction
the choices will probably fall on different lucky strategies, so the process
should be repeated several times, until stabilization.



\section{Greedy Buckets and the Virtual Best Strategy Selector}
\label{sec:vbs}

\begin{figure}
	\centering
	\includegraphics[scale=0.7]{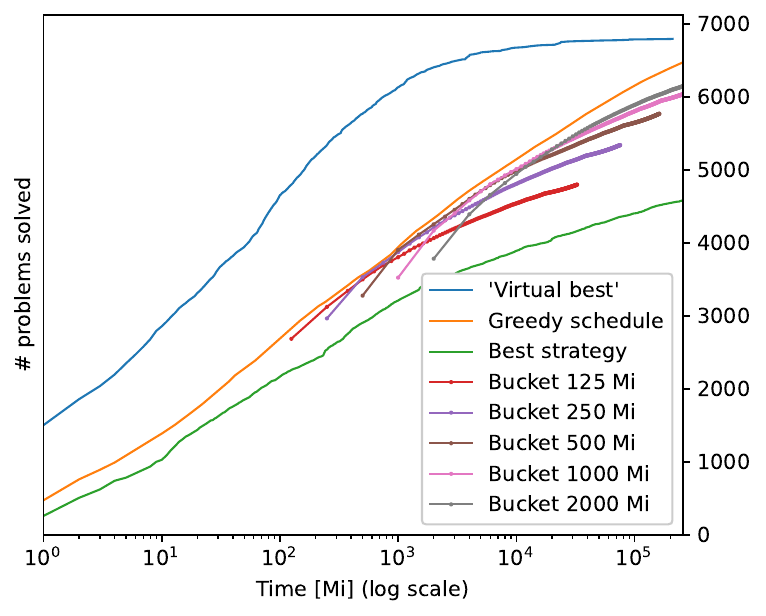}
	\caption{Cumulative performance of the main greedy schedule of Sect.~\ref{sec:casc}
		compared with greedily stacked equally sized strategy slices (buckets) of varying sizes.
		In addition, the performance of the virtual best strategy selector 
		and 
		the best strategy (at the \SI{256000}{Mi} mark) collected during strategy discovery. 
		$X$-axis clipped at \SI{256000}{Mi}.
	}
	\label{fig:vbs_append}
\end{figure}

A common way to experiment with the power of complementary strategies in a simple manner
is to evaluate each for a certain amount of time $t$
and then construct an equally sized schedule $(s_i,t)_{i=1}^n$ in a greedy fashion, starting 
with the best strategy and always adding the next one that brings the most problems not yet covered.
In Fig.~\ref{fig:vbs_append} we simulate such schedules for different time limits ($t=125,250,500,1000,2000$ Mi)
using our discovered strategies (of Sect.~\ref{sect:discover}) each time clipped at a respective value of $t$
(thus creating a $t$-sized ``bucket'' of problems).
There are two main points worth noticing. 
\begin{itemize}
	\item
	For all bucket sizes $t$, the bucketed schedule's performance starts well above the plotted best single strategy.
	This strategy $s_b$ was selected as the best performing one at the \SI{256000}{Mi} mark. For any of the bucket sizes $t$
	that we tried, there therefore exists in our pool a strategy that solves many more problems at $t$ than $s_b$.
	\item
	The bucket schedules never cross, but some come quite close to the performance of our ``free'' greedy schedule 
	(introduced in Sect.~\ref{sec:casc}), each, however, only at a specific time limit. Thus, for instance,
	just two greedy buckets of $t=500$ solve almost as many problems as our free schedule at \SI{1000}{Mi}.
\end{itemize}
Overall, however, we can see that the ability to schedule different strategies for different amounts of time is important 
for the overall performance of a schedule.

Fig.~\ref{fig:vbs_append} also plots the performance of the ``virtual best strategy selector'' (VBSS). This is a fictitious 
algorithm that, when given a problem, picks a strategy (from our pool) that performs the best on it.
The large gap between our greedy schedule and VBSS shows the opportunity of strategy selection approaches
(i.e., branching schedules), which we do not try to explore in this work. 

Another view of the VBSS plot is to consider
the $y$-axis as the main one: the plot then shows individual problems ordered (bottom to top) by the least amount
of time it takes some strategy of our pool to solve it. It is perhaps interesting to see that
\SI{94}{\percent} of all solvable problems (\num{6405} out of \num{6796}) can be solved (with some strategy)
under \SI{2000}{Mi}, i.e. roughly under 1 second (excluding parsing, c.f.~Sect.~\ref{sec:pdnc}). 

Finally, the VBSS plot also geometrically aligns with the plots of the bucketed schedules in an amusing way.
Reading VBSS at the $x$-coordinate of the schedule's plot's start, i.e., at $t$, we get the total amount 
of problems solvable under $t$, that is necessarily also the $y$-coordinate at which the bucketed schedule ends.


\section{Optimal (Pre-)Schedule Construction Encoding}
\label{sec:encodeSCP}

We want to split the given time budget $T$ among strategies $S$ in such
a way that the produced schedule solves as many problems as
possible. This can be formulated as a (mixed) integer programming
(MIP) task, e.g., as follows.

For every strategy $s\in S$, we have a non-negative variable
$t_s$ that represents the time allocated to $s$. These variables are
constrained by $\sum_{s \in S} t_s \leq T$. For every pair $s\in S$
and $p\in P$, we have a Boolean\footnote{In fact, we treat
these as integer variables with the upper bound 1.} variable $r_{s, p}$, where
$r_{s, p}=1$ means that strategy $s$ will claim a point for solving $p$.
Hence we require $r_{s, p}\cdot \SolveTimeP{s}{p}\leq t_s$ if $\SolveTimeP{s}{p} \neq \infty$, 
and otherwise set $r_{s, p}=0$. Moreover, we require that
$\sum_{s\in S} r_{s, p}\leq 1$, for every $p$, that is at most one
solution of a problem is rewarded. Finally, our criterion is to maximize $\sum_{s\in S, p\in P} r_{s, p}$.


\section{Strategy Gathering and Test Performance}
\label{sec:stamped}

\begin{figure}[t]
	\centering
	\includegraphics[scale=0.7]{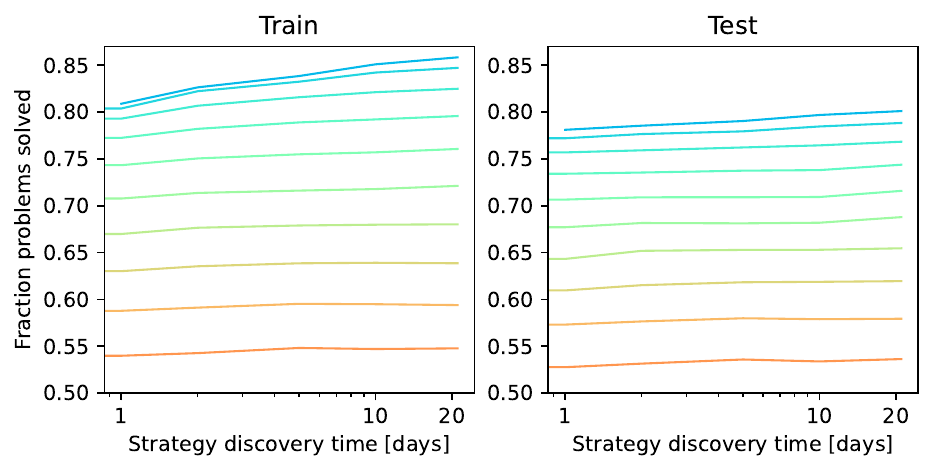}
	\caption{\emph{Train} and \emph{test} problems covered (as a fraction of available problems)
		by the base greedy algorithm, changing as a function of strategy discovery time (in days, log scale)
		of the strategies available. Colored lines connect budgets: \SI{2000}{Mi}, \SI{4000}{Mi}, \SI{8000}{Mi}, \ldots, \SI{1024000}{Mi}.
		Results averaged over 500 random 80:20 train/test splits.}
	\label{fig:stamped}
\end{figure}


Here we pause to estimate how the gathering of more and more strategies during the 
strategy discovery process affects performance on unseen problems. Fig.~\ref{fig:stamped}
shows, side by side, the train and test performance of the base greedy algorithm run
on increasingly larger subsets of our strategy set $S$. Each subset contains
only strategies discovered ``no later than'' a certain time stamp, where the time stamps
used determine the $x$ coordinate. Moreover, we plot the performance for several 
time budgets. The performance is averaged over many 80:20 train/test splits
and we maintain the methodology established in Sect.~\ref{sec:train_strategies_hygiene},
to always exclude those strategies whose witness problem is from the current test set.

What we can see is that (except for the budget of \SI{2000}{Mi} between days 5 and 10 where there is a slight deterioration)
even test performance always benefits from more training strategies being available to be utilized in a schedule.
For smaller budgets, however, the benefit from additional strategy discovery is small.


\section{When Does Problem-Picking Start?}
\label{sec:cherries}

\begin{figure}[t]
	\centering
	\includegraphics[scale=0.7]{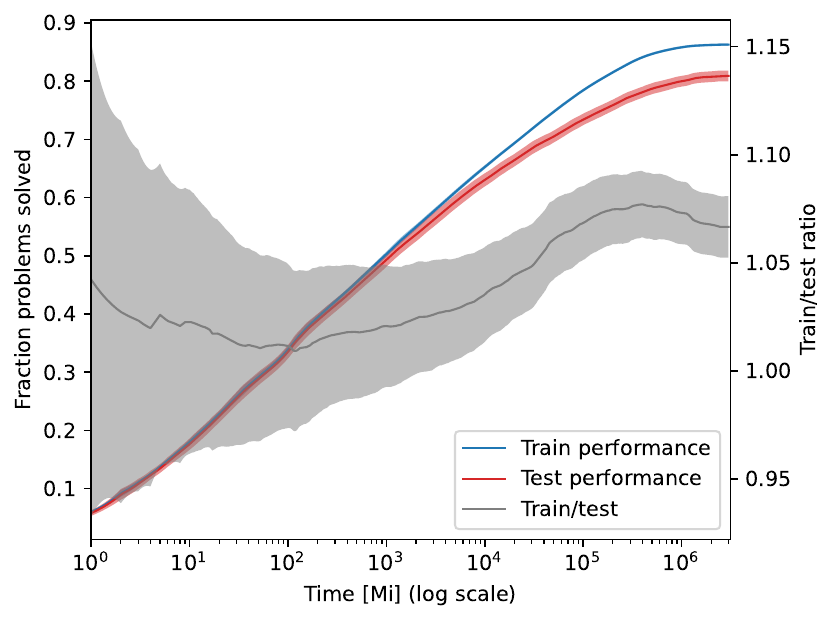}
	\caption{Average train and test performance (left $y$-axis) and their ratio (right $y$-axis) 
		of a base greedy schedule as a function of (anytime) budget.
		Shaded areas correspond to 1 standard deviation. Results averaged over 500 random 80:20 train/test splits.}
	\label{fig:cheeries}
\end{figure}


In Fig.~\ref{fig:cheeries} we plot the train and test performance and their ratio
of a base greedy schedule as a function of budget, averaged over 500 random 80:20 train/test splits.
The shaded areas highlight the corresponding variance (of $\pm 1$ standard deviation). 
Note that a higher variance of the test performance is mainly an artefact of rescaling to
``the number of available problems'', as the train set is 4 times larger than the test set.

The interesting part of the plot is the behavior of ratio between the train and the test performance .
The extremely high variance of this quantity up to approximately \SI{1e+3}{Mi}
points to a high influence of chance in the generalization capacity of the the greedy schedules there.
However, in these lower budgets we can expect good generalization on average.

As the budget increases, the ratio stabilizes in terms of variance but gradually
points towards poorer and poorer generalization. Our intuition is that, with increasing budgets,
covering additional problems from the training set 
becomes more and more an act of problem cherry-picking (training problem picking?) 
rather than an exploitation of a more fundamental strategy complementarity. 

Surprising to us was to see the final improvement of generalization in the regime between, approximately,
\SI{4e5}{Mi} and \SI{2e6}{Mi}, where the test performance begins to catch up.
A closer look at the data revealed that, just before this region, cherry-picking on the train set
would usually result in, e.g., 5 more training and 1 more testing problem covered per iteration. In this final
region of the graph, however, the schedule construction begins to claim individual training problems
per iteration, which---perhaps counterintuitively---often results in more than one test problem getting covered too. 


\section{Comparison of Our Greedy Algorithm with p-SETHEO}
\label{sec:psetheo}

p-SETHEO \cite{DBLP:conf/flairs/WolfL98} proposes a greedy pre-schedule construction algorithm.
Following the terminology of our greedy algorithm (see \cref{alg:greedy}),
p-SETHEO starts by initializing $t_s$ to 0 for all strategies.
Then, in each iteration, it extends $t_s$ by a the time step $\Delta T$
for a strategy $s$ that would solve the most additional problems by this extension.
In case no strategy solves any additional problems in $\Delta T$,
$\Delta T$ is increased by $\frac{T}{l}$,
where $l$ is the quantization parameter.

Our greedy algorithm does not rely on any quantization and works with arbitrary time granularity.
More importantly, for small $\Delta T$, p-SETHEO may waste schedule performance by being eager to solve relatively few problems early.
Our experiments with temporal reward adjustment (see \cref{sec:temporal_reward_adjustment,sec:experiment_regularization}) suggest that it is, in practice, beneficial to prefer relatively long slices that cover more problems.
On the other hand, increasing $\Delta T$ decreases the granularity and is likely to lead to assigning unnecessarily long limits to most strategies.

}

\newpage
\bibliography{main}

\end{document}